\documentclass[letterpaper, 10 pt, conference]{ieeeconf}  

\IEEEoverridecommandlockouts                   

\usepackage{cite}
\usepackage{amsmath,amssymb,amsfonts}
\usepackage{algorithmic}
\usepackage{graphicx}
\usepackage{textcomp}
\usepackage{xcolor}
\usepackage{hyperref}
\usepackage{makecell}
\usepackage{fancyvrb}
\usepackage{subcaption}

\title{\LARGE \bf
MRTA-Sim: A Modular Simulator for Multi-Robot Allocation, Planning, and Control in
Open-World Environments
}

\author{Victoria Marie Tuck$^{1\dagger}$, Hardik Parwana$^{2\dagger}$, Pei-Wei Chen$^{1}$, Georgios Fainekos$^{2}$, \\ Bardh Hoxha$^{2}$, Hideki Okamoto$^{2}$, S. Shankar Sastry$^{1}$, and Sanjit A. Seshia$^{1}$
\thanks{$^{\dagger}$This work was completed in part during an internship at Toyota Motor North America.}
\thanks{$^{1}$Department of Electrical Engineering and Computer Sciences, University of California at Berkeley, CA, USA
        {\tt\small \{victoria\_tuck, pwchen, shankar\_sastry, sseshia\}@berkeley.edu}}%
\thanks{$^{2}$Toyota Motor North America, MI, USA
        {\tt\small hardiksp@umich.edu, \{georgios.fainekos, bardh.hoxha, hideki.okamoto\}@toyota.com}}%
}

\begin{document}

\maketitle
\thispagestyle{empty}
\pagestyle{empty}

\begin{abstract}

This paper introduces MRTA-Sim, a Python/ROS2/Gazebo simulator for testing approaches to Multi-Robot Task Allocation (MRTA) problems on simulated robots in complex, indoor environments. Grid-based approaches to MRTA problems can be too restrictive for use in complex, dynamic environments such in warehouses, department stores, hospitals, etc. However, approaches that operate in free-space often operate at a layer of abstraction above the control and planning layers of a robot and make an assumption on approximate travel time between points of interest in the system. These abstractions can neglect the impact of the tight space and multi-agent interactions on the quality of the solution. Therefore, MRTA solutions should be tested with the navigation stacks of the robots in mind, taking into account robot planning, conflict avoidance between robots, and human interaction and avoidance. This tool connects the allocation output of MRTA solvers to individual robot planning using the NAV2 stack and local, centralized multi-robot deconfliction using Control Barrier Function-Quadrtic Programs (CBF-QPs), creating a platform closer to real-world operation for more comprehensive testing of these approaches. The simulation architecture is modular so that users can swap out methods at different levels of the stack. We show the use of our system with a Satisfiability Modulo Theories (SMT)-based approach to dynamic MRTA on a fleet of indoor delivery robots.

\end{abstract}



\section{Introduction}

\begin{figure}
    \centering
    \includegraphics[width=\linewidth]{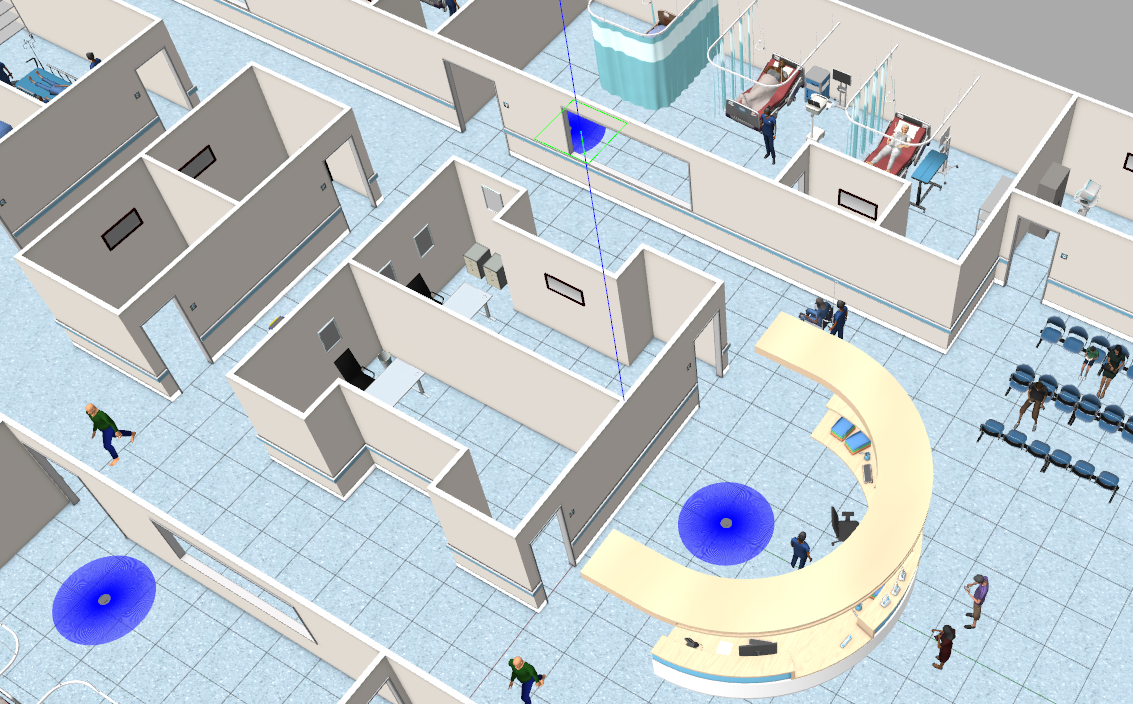}
    \caption{View of robots (center of the blue circles) moving in the simulation. Hospital beds, crowded rooms, humans, and tight doorways create difficult movement challenges for multi-agent, continuous operation.}
    \label{fig:main-pic}
\end{figure}

\begin{figure*}[!t]
    \centering
    \includegraphics[width=\linewidth]{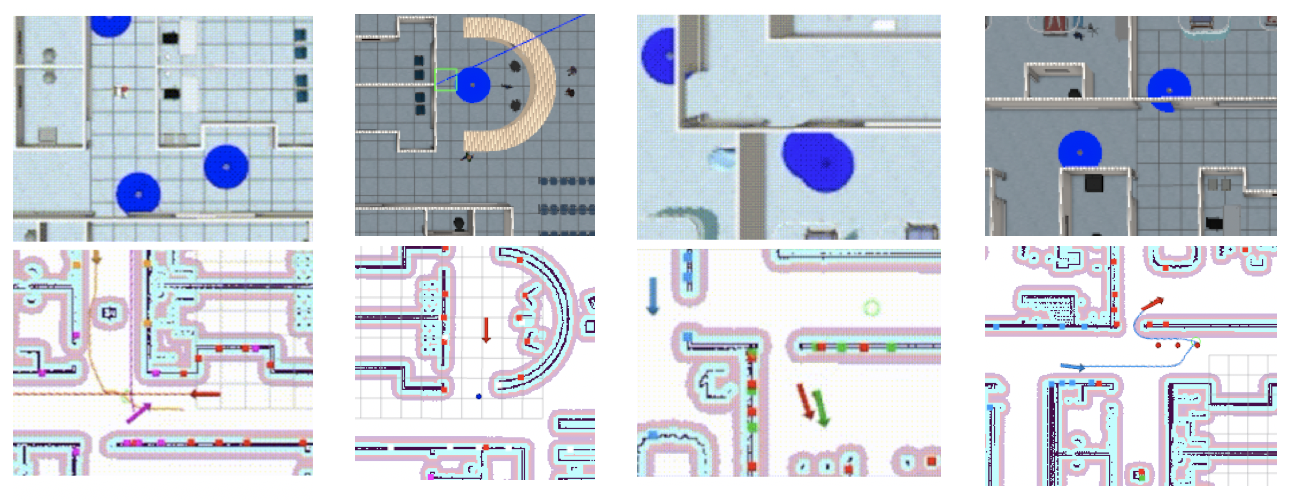}
    \caption{The Gazebo view is shown in the top row; the bottom row shows the Rviz visualization of paths, obstacle points, and agent positions and headings. From left to right: a) The yellow and pink agents follow the "right-hand" travel rule on the roads. The red agent is waiting for the higher priority pink agent to pass first due to the CBF cluster control b) A moving human is shown in Gazebo and Rviz. c) Due to CBF cluster control, the green agent has moved out of the way of the red agent that just arrived. d) The blue agent originally plans for the end of the queue (the right-most small dot) before it knows its queue position. Once it receives its position, it adjusts its path to plan towards the room. The red agent is not in the queue because it has completed its tasks.}
    \label{fig:sim_stills}
\end{figure*}

Centralized multi-robot systems are being used in many new areas such as advanced air mobility, autonomous vehicles, environmental monitoring, disaster management, and hospital robotics \cite{GARROW2021103377} \cite{JeonLK2017} \cite{BAYAT201776} \cite{NABAVI2022689}. In industries such as healthcare, robots can be used to deliver medications and supplies, provide a video platform for a call to family, or help with rehabilitation \cite{morgan2022robots} \cite{yuan2021systematic}. Such multi-robot systems are already being used in warehouses \cite{WurmanDAM2008aimag}. 
However, these environments are strongly structured and largely predictable. 
In contrast, novel applications involve open-world environments and unpredictable elements such as humans.

Multi-robot systems are often developed with a hierarchical approach where tasks are assigned at the highest level, individual robots plan and follow trajectories, and groups of robots must implement deconfliction and deadlock-reducing strategies. 
In structured environments like warehouses, we see these multi-robot systems being successfully used. 
However, when moving out of the warehouse into more open-world environments, it can be difficult to analyze such environments theoretically making the efficacy of these approaches less clear. 
As real-world testing can be costly, digital twins for testing in simulation can be used for realistic evaluation faster and at a lower cost. 
Another major challenge when comparing approaches is the lack of methods implemented on a common platform for comparison. 
This need has been highlighted in the validation section of the recent Road map for Control for Societal-Scale Challenges \cite{Annaswamy2023}. 
Finally, algorithms for robot movement and interaction are often compared only on short, isolated tasks -- such as contrasting a control algorithm against another for a two robot avoidance case. 
Yet, in practical applications, these algorithms must operate continuously within a full system over extended periods. 
Therefore, it is crucial to evaluate methods on longer tasks and as part of a complete algorithm stack to better understand their real-world performance.

In this work, we introduce MRTA-Sim, a simulation platform for testing approaches to Multi-Robot Task Allocation Systems (MRTA) in open-world environments. We build upon Gazebo Classic, an
open-source robotics simulator and use the Robot Operating System 2 (ROS2) for robot operation and coordination \cite{gazebo} \cite{ros2}. Our simulator fulfills the need for a common platform on which to evaluate not only task allocation approaches with path planning and control approaches but also different path planning and control approaches when implemented for long-term multi-agent systems. We include a baseline task allocation, path planning, and control algorithm as well as infrastructure to coordination many agents at different levels of centralization. Visualization tools are included to understand the movements of the agents.

The specific contributions of this work are
\begin{enumerate}
    \item the MRTA-Sim shown in Fig. \ref{fig:main-pic} and Fig. \ref{fig:sim_stills} which provides a challenging and unpredictable environment for benchmarking of MRTA, path planning, and control approaches on long-term multi-robot multi-task problems;
    \item a set of baseline algorithm implementations so that users can more easily evaluate their new approach 1) with respect to a turnkey baseline and 2) within a system that must also complete task allocation, path planning, or control; 
    \item an interface to the scenario generation tool Scenic~\cite{Fremont_2019} for creation of different settings; and
    \item an analysis of scalability of the system and its base controller.
\end{enumerate}

We describe related simulation and hardware platforms in Sec. \ref{sec:related_work}. In Sec. \ref{sec:mrta_sim}, we introduce the tool's use case, architecture, and multi-agent interaction methods. We provide brief usage instructions in Sec. \ref{sec:usage}, and Sec. \ref{sec:experiments} includes scalability experiments.

\section{Related Work \label{sec:related_work}}

\begin{table*}[ht]
    \centering
    \caption{Comparison of Robotic Simulators and Hardware Platforms}
    \begin{tabular}{|c|c|c|c|c|c|c|c|c|c|}
        \hline
         & \multicolumn{3}{|c|}{Robot System Layers} & \multicolumn{3}{|c|}{Robot Coordination} & \multicolumn{3}{|c|}{Environment} \\
         \hline
        \thead{\textbf{Features $\rightarrow$} \\ \textbf{$\downarrow $ Simulator}} & \textbf{Control} & \thead{\textbf{Path} \\ \textbf{Planning}} & \thead{\textbf{Task} \\ \textbf{Allocation}} & \thead{\textbf{Multi-Robot:} \\ \textbf{Centralized}} & \thead{\textbf{Multi-Robot:} \\ \textbf{Local}} & \thead{\textbf{Multi-Robot:} \\ \textbf{Decentralized}} &  \textbf{Humans} & \thead{\textbf{Physics-} \\ \textbf{based}} & \thead{\textbf{Real} \\ \textbf{Hardware}}  \\ \hline
        \textbf{Robotarium \cite{8960572}} & \checkmark & & & & & & & \checkmark & \checkmark \\ \hline
        \textbf{ARGoS \cite{Pinciroli:SI2012}} & \checkmark & & & & & \checkmark & & \checkmark & \\ \hline
        \textbf{Arena \cite{arena4}} & \checkmark & \checkmark & & & & & \checkmark & \checkmark &  \\ \hline
        \textbf{Open-RMF \cite{openrmf}} & \checkmark & \checkmark & \checkmark & \checkmark & & & \checkmark & \checkmark & \\ \hline
        \textbf{VMAS \cite{bettini2022vmas}} & \checkmark & & & \checkmark & & \checkmark & & &  \\ \hline
        \textbf{REMROC\cite{heuer2024}} & \checkmark & & & \checkmark & & \checkmark & \checkmark & \checkmark & \\
        \hline
        \textbf{MRTA-Sim [ours]} & \checkmark & \checkmark & \checkmark & \checkmark & \checkmark & \checkmark & \checkmark & \checkmark &  \\ \hline
    \end{tabular}
    \label{tab:sim_comparison}
\end{table*}

We provide a brief overview of related simulation platforms for robotic evaluation.
A summary of related work is found in Table \ref{tab:sim_comparison}.
The first portion of this table looks at whether or not the platform supports algorithms at each level of a multi-robot system including control, path planning, and task allocation.
The second portion considers multi-robot problems.
We label a platform \textit{Multi-Robot:Centralized} if it has a centralized coordination system between the agents.
If agents have access to local position information of other agents that they use to avoid conflicts, we assign the label \textit{Multi-Robot:Decentralized}.
Lastly, an intermediate notion of coordination (\textit{Multi-Robot: Local}) indicates that agents can i) determine leaders among locally located agents for the leaders to ii) determine control for all local agents.
We describe if the platform's environments have realistic physics modeling (\textit{Physics-based}) and/or modeled humans (\textit{Humans}). 
If the platform is a hardware platform (e.g., remote-access robots), we label it as \textit{Physics-based} and also include it under \textit{Real Hardware}.

We compare platforms that are remote-access robot hardware systems or benchmarking simulators for mobile robotics. We acknowledge other important simulators such as Maniskill, robosuite, and Isaac Sim but exclude them from this comparison as they are focused more on completing single robot tasks with an emphasis on scalable training for learning \cite{tao2024maniskill3gpuparallelizedrobotics} \cite{robosuite2020} \cite{NVIDIAIsaacSim}. The Robotarium provides a platform for multi-agent testing on real robots and provides a safety control layer but has little support for multi-layered architectures and only obstacle projection \cite{8960572}. 
ARGoS handles large numbers of agents but also only includes a controller plugin without the option for higher levels of planning including path planning and task allocation \cite{Pinciroli:SI2012}. 
VMAS also handles large numbers of agents but is focused on testing and training end-to-end learning approaches without consideration for hierarchical systems and supports only simplistic Python environments \cite{bettini2022vmas}. 
Arena offers complex environments that include dynamic components but currently has no support for multi-agent settings \cite{arena4}. 
Open-RMF is a tool that considers more of the hierarchical aspects with task allocation and deconfliction \cite{openrmf}. 
However, it handles agent deconfliction via use of a traffic schedule, which is highly centralized and potentially less robust to environment changes. 
Remroc is a simulator in Gazebo that tests multi-agent systems but with a focus on multi-agent path finding problems and multi-agent coordination without the task allocation component \cite{heuer2024}. Particularly important for robotic applications, it lacks the typical robot path planning level. Across all of these options, we see a need for a system that both handles multi-robot coordination at various levels of centrality and can be used for longer-term operation with online arriving tasks. 

\begin{figure*}
    \centering
    \includegraphics[width=0.9\linewidth]{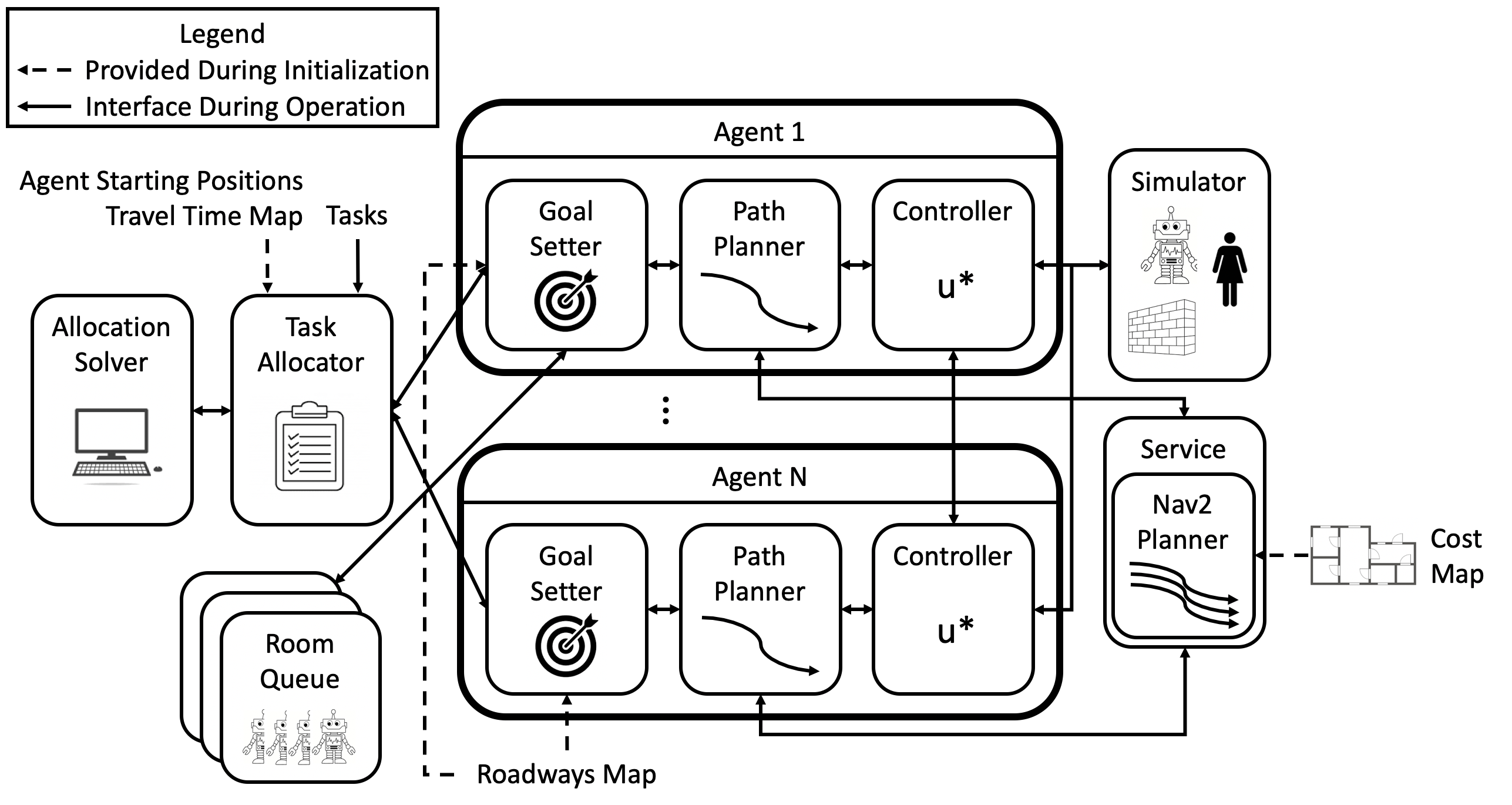}
    \caption{MRTA-Sim Architecture}
    \label{fig:architecture}
\end{figure*}

\section{Multi-Agent Task Allocation SIMulator (MRTA-SIM) \label{sec:mrta_sim}}

In this section, we provide an overview of the components of the tool, images of which is shown in Fig. \ref{fig:sim_stills} and the architecture of which is shown in Fig. \ref{fig:architecture}. The tool can be accessed at \href{https://github.com/victoria-tuck/multi-robot-task-allocation-stack}{https://github.com/victoria-tuck/multi-robot-task-allocation-stack}. Section \ref{sec:architecture} explains each component of the system. Section \ref{sec:multi-agent} goes into detail about how robots deconflict and avoid deadlocks.

We use the notation $[N] = {1, \ldots, N}$ to denote the set of integers from 1 to N. The notation $S_{-i}=S  \setminus i$ where $S$ is a set denotes the set with element $i$ removed.

\subsection{Problem and Use Case}

Robots in complex, real-world environments need to be able to respond and adjust to changing environments. 
These systems should operate autonomously for long periods of time and, therefore, need robust task allocation, planning, and control algorithms. 
Operating in open-world scenarios is hard to analyze theoretically, which is where realistic physics-based simulations (digital twins) help validate approaches for real-world use. 
However, no open source physics-based simulator currently provides a turnkey multi-agent system that includes features such as task allocation, decentralized planning, control, and deconfliction.
This means researchers lack a platform to test their approaches across these levels within the context of a long-running task allocation and planning system.
Our goal is to create such a simulator that allows researchers to plug their new approach into the system at any level, leveraging the baselines provided and implementations of other algorithms in the stack to test on sequences of tasks in a complex environment.

\subsection{Architecture \label{sec:architecture}}

In this section, we provide an overview of the architecture of the tool. Figure \ref{fig:architecture} provides a complete view of the system, and each block is explained in detail below.

\subsubsection{Allocation Solver}

The allocation solver determines which agents complete which tasks and in what order. New tasks are received from the task allocator as well as feedback on which tasks have been completed by what time by each robot. The Allocation Solver has access to travel times between system locations and agent starting rooms. It needs to either directly output a sequence of system locations for each agent to track or needs to output this information in a form factor that the user interprets in the task allocator to create the sequences. The system locations $V \subset P$ are given in the form of a travel time graph $G = (V, E)$. We denote the set of valid robot positions as $P$. Additionally, if the user would like to loosen the need for travel time information, they can adjust the function of the task allocator and provide a new solver while still using the rest of the system.

We have tested the SMrTa solver and specless for task allocation in our framework \cite{tuck2024smt} \cite{specless}. The SMrTa solver is a Satisfiability Modulo Theories (SMT)-based approach to Multi-Robot Task Allocation, and the Python library SPECification LEarning and Strategy Synthesis (SPECLESS) is a tool for learning specifications from demonstrations and synthesizing strategies. Both of these require knowledge of approximate travel times between system locations.

\subsubsection{Task Allocator}

There is a single task allocator, which calls the allocation solver when a new set of tasks arrives, sends a sequence of waypoints to the waypoint generator for each agent, and takes feedback from the waypoint generator on actual waypoint arrival times. 

The solver is initialized with a fully connected, weighted, undirected graph of travel times $G = (V, E)$ where an edge $e_{i,j} = (v_i, v_j) \in E$ connects two locations of the system $v_i \in V$ and $v_j \in V$ with a weight $w_{ij} \in \mathbb{R}_{+}$ that is an approximation of the travel time between locations $v_i$ and $v_j$. 
The task allocator currently supports connection to the SMrTa solver \cite{tuck2024smt}.
The task allocator calls the allocation solver on the set of robots $\mathcal{N}$, set of incoming tasks $\mathcal{M}_j$, and travel time graph $G$ and outputs a sequence of waypoints $W_{i}=w_{i,1}, \ldots w_{i,M}, \ \forall i \in [N]$ where each waypoint $w_{i,m} \in W_{i} $ is a valid position $\in P$. The sequence of waypoints may change whenever new tasks arrive to the system.

Due to the constrained and complex nature of the environment, robots may not always take the expected time to complete tasks. 
Therefore, the task allocator receives actual travel time information as a sequence of the times that each robot actually arrived at each waypoint. This sequence of times is sent as feedback back to the waypoint generator as $t_{i, j}, \ \forall w_{i,j} \in W_i, \ \forall i \in [N]$.

\subsubsection{Room Queues}

Room queues are used to assist in deadlock reduction between agents and are described in more detail in Sec. \ref{sec:multi-agent-queue}. 
Queues are established for each room or heavily congested areas. 
A robot's waypoint generator plans a path to the end of the queue and requests a spot in the queue once nearby. 
The waypoint generator updates the sequence of waypoints based on the returned queue position (e.g., if a robot initially planned toward some waypoint $w \in P$ that is the location of the final position of the queue but is placed at the top of the queue, the waypoint generator will add onto the path to move into the room). 
Only the robot at the head of the queue is allowed in the room and will hold onto the position until it has moved a certain distance away from the room or finished all tasks. Once it finishes, the queue updates and the next robot receives access. 

\subsubsection{Waypoint Generator}

Each robot has three components - a waypoint generator, path planner, and controller. The highest level component, the waypoint generator, takes in a sequence of high level actions to complete from the task allocator, converts each action (such as "pick up object A") into a sequence of waypoints and continuously provides the next two waypoints (the one the agent is currently heading to and the following) to the planner. The waypoint generator will follow the roadways and manage requesting room access via the room queues as well as update the waypoint sequence based on the room queue position. At minimum, the roadways include the starting and ending point of each route is specified as the path to travel between two system locations $l_i, l_j \in L$. Intermediate waypoints to designate the paths to travel on the roadways may also be specified. The waypoint generator monitors the actual time that each point is reached and stores that for feedback to return to the task allocator.

\subsubsection{Path Planner}
The path planner calls the planner service on the waypoints received from the waypoint generator  to determine a path for the robot. 
The robot waits to receive valid plans from the planner before proceeding.

\subsubsection{Controller} \label{sec:controller}

The controller has access to the robot's state, goal location, and size; the other robots' states; nearby obstacles; and the humans' states and velocities.
For computational reasons, the nearest obstacles are found via ray tracing along pre-specified directions about the robot.
The nearest obstacle point along the ray is captured and represented as a list of obstacle points to the robot.

In the current base controller for the included turtlebot robot, we support the unicycle dynamics
\begin{equation*}
    \begin{aligned}
    \dot{x} &= v \cos \theta, \\
    \dot{y} &= v \sin \theta, \\
    \dot{\theta} &= \omega.
    \end{aligned}
    \label{eq:unicycle}
\end{equation*}
with position $x$ and heading $\theta$ as the state and angular velocity $\omega$ and velocity $v$ as the control. Other robots with unicycle dynamics  can be exchanged in the world file and tuned in the controller. If different dynamics or robot state are needed, these can be changed in the controller wrapper, and the user would need to provide a controller for their desired dynamics or provide a dynamic representation to the given controller module.
The base controller used is a CBF-QP, which we include because of the current interest in CBFs in the controls community \cite{ames2019control}. Control is determined for navigation to the closest point on the planner's path at least a minimum distance $\delta$ away; this parameter is currently set to 0.5m but can be changed in the controller. An individual CBF-QP is used when far away from others and a local, centralized CBF-QP is used when near other active robots. As explained in more detail in Section \ref{sec:local_cbfqps}, the wrapper around a robot's controller determines their cluster for the local, centralized CBF-QP. 
Control is computed at a rate of 20 Hz but can be adjusted based on desired rate. 

\subsubsection{Simulator}

\begin{figure}
    \centering
    \begin{subfigure}[b]{0.51\linewidth}
        \includegraphics[width=\linewidth]{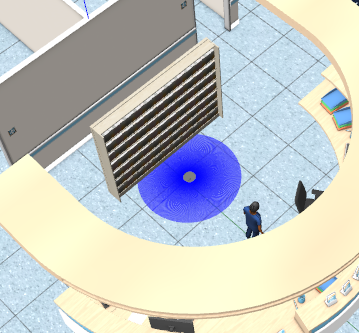}
    \end{subfigure}
    \hfill
    \begin{subfigure}[b]{0.45\linewidth}
        \includegraphics[width=\linewidth]{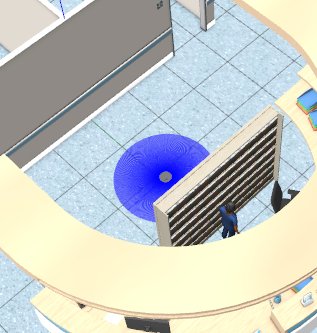}
    \end{subfigure}
    \caption{The bookshelf in these two screenshots has been added via a Scenic program. The distance away from the wall for each was automatically sampled from a distribution by Scenic.}
    \label{fig:scenic}
\end{figure}

For simulation, we use Gazebo Classic, an open-source robotics simulator.
Gazebo requires a world file for environment description; we include the AWS RoboMaker Hospital World as the base environment. The interface was written and tested in ROS2 Humble. 
Humans track waypoints that can be specified in a yaml file; they are modeled using the popular social force model \cite{social_force}. We also provide a basic interface to Scenic, a probabilistic programming language for scenario generation \cite{Fremont_2019}. Figure \ref{fig:scenic} shows a bookshelf that has been added in a randomly sampled position by a scenic program. We note that the currently provided costmap for obstacle avoidance does not take Scenic-generated objects into account, so these will need to be added into the costmap by the user.

\subsubsection{Planning Service}

The planning service requires a waypoint and current position as ROS2 poses and returns a Nav2 Path object. We use a single Nav2 stack with a wrapper node, which all robots call for planning \cite{macenski2020marathon2}. 
This wrapper node allows a single Nav2 stack to be used as a planning service so a single computer can support simulating navigation for large numbers of robots at once. 
The base planner finds paths using a cost-aware 2D-A$^*$ planner \cite{macenski2024nav2} on a map of known static obstacle locations with a pre-computed cost map.

\subsection{Multi-Agent Interaction \label{sec:multi-agent}}

In this section, we provide further information about the provided approaches for conflict avoidance and deadlock reduction. 
We use common approaches found in the literature as baselines for comparison in future research.

\begin{figure*}
    \centering
    \includegraphics[width=0.9\linewidth]{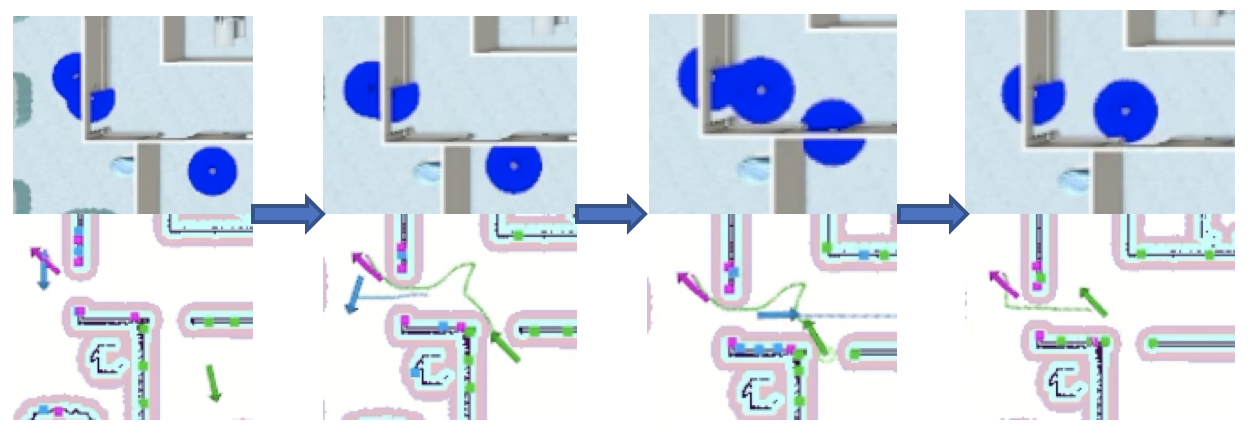}
    \caption{From left to right: All agents are stopped because they have no active tasks. Then the blue and green agents received tasks. The green agent waits for the higher priority blue agent to pass. Then, it replans towards its goal and moves to follow its path.}
    \label{fig:example}
\end{figure*}

\subsubsection{Roadways} \label{sec:multi-agent-roadway}

A network of road-like paths has been designed for the floor so that agents can more easily navigate past each other in corridors. Paths follow a right-hand rule and are described via sequences of waypoints. The waypoint generator will pull waypoints to follow from the network of roads.

\subsubsection{Local, Centralized Control via Control Barrier Function Quadratic Programs (CBF-QPs) \label{sec:local_cbfqps}}

When a robot is alone, it calculates its control via a CBF-QP with slack using a distance based CBF to avoid obstacles \cite{ames2019control}.
For deconfliction and deadlock avoidance in the presence of other robots, a local, centralized CBF calculation among clusters of robots is used to compute the control for each robot in the cluster.
For each robot $i$ to determine its cluster, it first uses the positional information of other agents $p_{j \in [N]_{-i}}$, to determine its set of neighbors $B_i = \{ j \in [N]_{-i} \ | \ ||p_j - p_i|| < d_{min} \}$.
Each robot shares $B_i$ and creates an undirected graph $G_c = (\mathcal{N}_c, \mathcal{E}_c)$ where an edge in the graph indicates two agents that are neighbors. The graph is formed with nodes $\mathcal{N}_c = [N]$ and edge set $\mathcal{E}_c = \{ (i, j) \ | \ j \in B_i, i \in [N] \}$.
Let $\mathcal{T} = {C_1, \ldots, C_k}$ be the set of clusters (isolated components) in the cluster graph resulting from taking the transitive closure of the connection graph $G_c$. Each $C_i \in \mathcal{T}$ is a set of agents that are in a cluster. Cluster vertices are disjoint ($C_i \cap C_j = \emptyset \quad \forall i \neq j = 1, \ldots k$) as agents can only belong to one cluster.
The active robots of the cluster are the robots with tasks currently assigned that are not waiting in a queue.
The leader of each cluster is a single agent determined by priority assignment out of the active robots. We pre-designate a priority ordering among agents that is used throughout the simulation, but an alternative approach is to determine the leader based on another metric such as distance to goal such as in Garg et al. \cite{10886421}.
Each robot updates its current cluster and cluster leader at a rate of 20 Hz.

The leader calculates control for all agents in the cluster and sends remote control commands to the other cluster agents.
We calculate the distance $d_i$ an agent $i$ is from their next waypoint, the desired heading angle $\theta^*_i$ that points directly towards the waypoint $w_i$  from their current position $p_i$ and the heading angle error $\hat{\theta_i}$ with respect to their current heading $\theta_{i}$ as
\begin{align*}
    d_i &= ||p_i - w_i||_2 \\
    \theta_i^* &= \arctan2(w_{i,y} - p_{i,y}, w_{i,x} - p_{i,x}) \\
    \hat{\theta_i} &= \theta_i^* - \theta_i \\
    v_i^* &= \text{min}(k_vd_i\cos(\hat{\theta_i}), v_{max}). 
\end{align*}
In our CBF-QP implementation and found in \textit{multi\_dynamic\_unicycle.py}, the leader $i$'s nominal control $u_i^*$ towards waypoint $w_i$ from their current position $p_i$ with velocity $v_i$ is their desired control
\begin{equation*}
    u_i^* = 
\begin{cases}
    [0, k_{\theta}\hat{\theta_i}]^T  & \text{if } \hat{\theta_i} < \bar{\theta}\\
    [k_v (v_i^* - v_i), k_{\theta}\hat{\theta_i}]^T,              & \text{otherwise},
\end{cases}
\end{equation*}
and the nominal control $u_j*$ for the $j$th other agent in the cluster is to slow to a stop

\begin{equation*}
    u_j^* = [\text{min}(\text{max}( k_{slow} v_j,-v_{max}), v_{max}), 0]^T.
\end{equation*}
The tuning parameters are $k_v, k_{\theta}, and k_{slow}$.
Note that agents stop moving towards their waypoint once they are less than a minimum distance $d_{min}$ away. Using local clusters allows for scalable centralized multi-agent deconfliction, and the remote control approach assists in deadlock avoidance.

\subsubsection{Queueing \label{sec:multi-agent-queue}}

Each room or congested area with a system location has a queue to manage access.
The queue receives requests to add robots and will remove a robot when signaled to by the robot when the robot has exited the area.
An agent's queue requests are managed by the waypoint generator.
Queue positions are placed outside the rooms.
With queueing, robots are less likely to get deadlocked at doorways and within rooms.
However, if this wait time is not properly accounted for in the task allocator, this can cause a mismatch between expected and actual task completion time.

\section{Tool Usage Instructions \label{sec:usage}}

Section \ref{sec:travel_time} explains the tool we provide to collect travel time information for use within the task allocator.
Section \ref{sec:get_started} contains more information about how to get started. An example portion of a run is shown in Fig. \ref{fig:example}

\subsection{Travel Time Collection \label{sec:travel_time}}

In many approaches, we need an estimate of travel time between relevant locations.
We provide a script to move a single agent between all pairs of locations in the system to collect travel time information.
Currently, we use the maximum time for each pair in the travel time graph for the task allocator though this can be changed in the travel time script if another metric like an average is preferred.

\subsection{Getting Started \label{sec:get_started}}

We provide a brief overview here and direct the reader to our github repository \href{https://github.com/victoria-tuck/multi-robot-task-allocation-stack}{https://github.com/victoria-tuck/multi-robot-task-allocation-stack} for further instructions on tool use.
Requirements include a dedicated NVIDIA GPU for graphics and a Linux install with docker.

A user specifies the starting locations of the robots in the robot setup file and case config file as shown in Fig. \ref{fig:config}.
A testcase file describes the sequence of tasks to arrive to the system.
Fig. \ref{fig:tasks} shows one such task request which would be included as one element of the system's task stream.

\begin{figure}[b]
\centering
    \begin{Verbatim}[fontsize=\small, samepage=true, commandchars=\\\{\}]
    agents:
      robot1:
        start: [0, 2.2]
      robot2:
        start: [4.25, -27.2]
    \end{Verbatim}
    \caption{Example configuration file for two agents with starting positions $(0, 2.2)$ and $(4.25, -27.2)$.}
    \label{fig:config}
\end{figure}

\begin{figure}
    \begin{Verbatim}[fontsize=\small, samepage=true, commandchars=\\\{\}]
    "arrival": 40,
    "tasks": [
      \{"start": 3, "end": 0, "deadline": 150\},
      \{"start": 2, "end": 1, "deadline": 300\}
    ]
    \end{Verbatim}
    \caption{Example of how to specify a task request to the system with the SMrTa solver. The two tasks arrive at t=40. The first requires a pickup from room 3, a drop-off in room 0, and must be completed by t=150. The second has a similar requirement.}
    \label{fig:tasks}
\end{figure}

\begin{figure}
    \centering
    \includegraphics[width=0.8\linewidth]{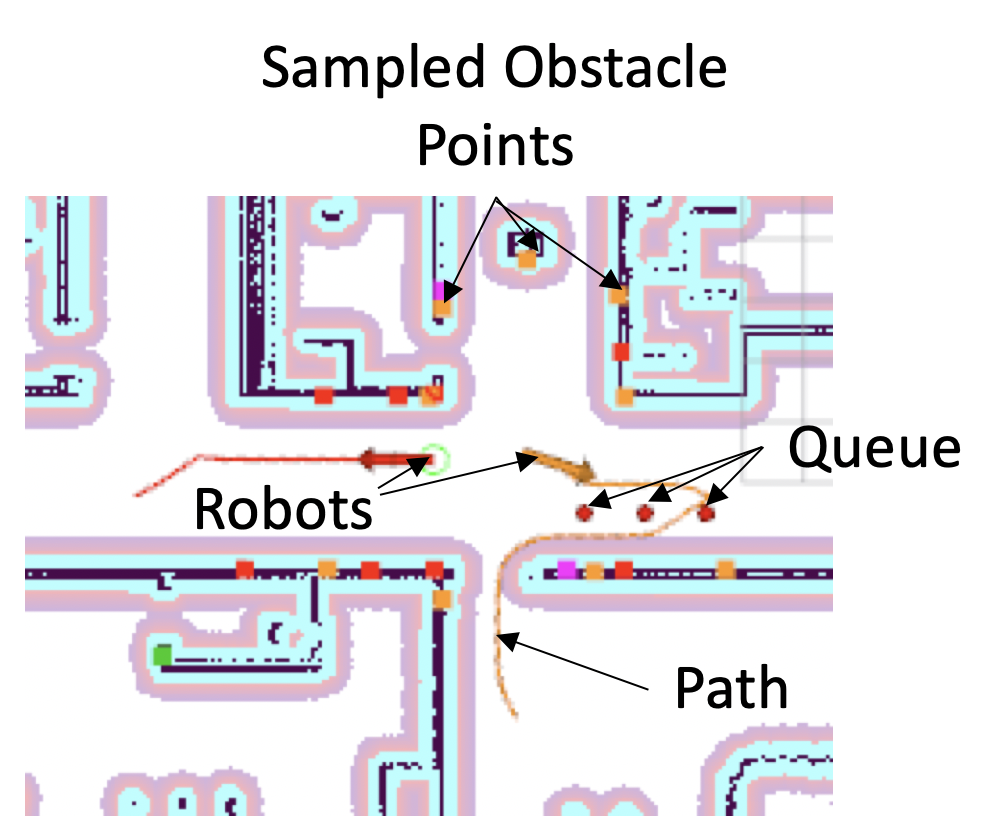}
    \caption{Agents, a queue outside a room, the sample obstacle points, and the agent paths are shown in Rviz. Walls are shown in black, and the blue and pink regions show the costmap for obstacle collision avoidance.}
    \label{fig:rviz}
\end{figure}

Setup is simplified by our use of docker. Relevant dependancies for docker and nvidia-docker should be installed, and a dockerfile is included to set up the environment.
Because we use ROS, the colcon workspace needs to be built; instructions for this and other steps can be found in our README. The testcase file is specified within the task allocator node (\textit{dispatcher.py}).
Additionally, we include many aliases to shorten terminal commands to start Gazebo, Rviz, and different nodes. These can be found in the README. We have an Rviz visualization to better understand the movements of the robots shown in Fig. \ref{fig:rviz}. 

\begin{figure}
    \centering
    \includegraphics[width=0.7\linewidth]{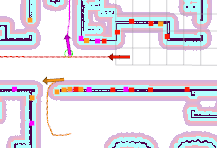}
    \caption{Showing three agents in Rviz. Agents and their current direction are represented by the arrows, and the lines of corresponding color are their planned paths.}
    \label{fig:planning}
\end{figure}

In Fig. \ref{fig:planning}, we show the planned paths of three agents near each other.
The bottom (yellow) agent has received access to the bottom room so is traveling there.
It is close enough to the other agents, so all three have formed a cluster with the bottom agent as the leader.
The other agents are paused due to the cluster, and the upper (pink) one will move once the bottom one moves further away as it becomes leader of the smaller cluster.
This is a screenshot from a simulation with six agents receiving 6 pick and drop tasks total in two sets.
In this run, two agents do run into planner issues and therefore do not complete their tasks, which is unsurprising given that planners may fail in certain cases.
However, the use of our tool will make easier designing and testing approaches to such planning and other issues like deadlocks.

The user can swap out approaches by implementing their approach as a new function and replacing the current function call to  the controller, allocation solver, or planning approach.
The planning approach can be changed inside the planner service or can replace the planner service.
The CBF-QP function call can be swapped in the controller block, and the task allocator similarly has an allocation solver function call that can be adjusted.
An additional swap that can be made is to change the prioritization when determining the cluster leader. 
Swapping allows the user to not only test the component in a environment but to test it within the context of the rest of the system. In the future, we intend to replace the manual swap in the code with a configuration file where the user can specify the methods to be used.

\section{Experiments \label{sec:experiments}}

In this section, we study the scalability of our system.
All experiments are run on a computer running Ubuntu 20.04 with 20 3.5GHz Intel Cores i9-9900X and 64GB RAM and a dedicated Nvidia GPU for Gazebo graphics.
In Table \ref{table:compute_times}, we analyze the time to compute computationally heavy calculations for control on a six agent example.
The single-agent CBF-QP is invoked when an agent is far away from others and can run at up to 100 Hz if desired.
The multi-agent CBF-QPs are invoked when in a cluster of that size, and each multi-agent CBF-QP is solved by only one agent of the cluster.
The initial QP time refers to the time that it takes to run the first call of this QP.
We use JAX for gradient calculation in the constraints of the QPs, and this time difference is due to the JITing required for the first call \cite{frostig2018compiling}. These results show us that the QP-based control can run in real time (\>20 Hz) as the initial call can be run before the run starts.

\begin{table}[h!]
\centering
\caption{Average Time for CBF-QP Initial and Other Control Calculations (s)}
\begin{tabular}{|c|c|c|c|c|}
\hline
 & \textbf{1-Agent} & \textbf{2-Agent} & \textbf{3-Agent}  \\ \hline
 Init & 0.0126 & 1.50 & 1.54 \\
 \hline
Other & 0.00854 & 0.0143 & 0.0123 \\ \hline
\end{tabular}
\label{table:compute_times}
\end{table}

We also study the scalability by looking at the real-time factor with respect to 2, 4, and 6 robotic agents in the simulation in Table \ref{table:scalability}. The table includes the lowest real-time factor seen when running the simulation. Physics-based simulators can struggle to handle too many agents at once, which we do see in our simulator. In the future, we wish to switch to a setup across multiple computers to mitigate this issue.

\begin{table}[h!]
\centering
\caption{Real-Time Factor}
\begin{tabular}{|c|c|c|c|c|}
\hline
 \textbf{Number of Agents} & \textbf{2} & \textbf{4} & \textbf{6}  \\ \hline
 Real-Time Factor & 0.73 & 0.56 & 0.42 \\ \hline
\end{tabular}
\label{table:scalability}
\end{table}




\addtolength{\textheight}{-3cm}

\section{Conclusion}

In this work we introduce the MRTA-SIM tool, an open source tool for testing long-term multi-robot task allocation systems with hierarchical robot stacks in a realistic physics simulator.
Users of the tool can test their Multi-Robot Task Allocation (MRTA) approach with standard deconfliction, planning, and control approaches and/or study how different deconfliction, planning, or control approaches affect the success of the MRTA approach in this complex setting. Additionally, our implementation supports multi-agent coordination algorithms with different levels of centrality.
We note that robot-to-robot deconfliction and resolving deadlocks are not solved problems, especially when such systems are deployed in unstructured spaces.
Therefore, although we implement methods to mitigate these issues, our system is not completely free of these issues, leaving room for this system to be used to develop better approaches.
Future directions include adding intermittent corridor blockages, expanding the environments supported, and modeling communication outages. We also intend to add increased support for the Scenic interface and improve accessibility and scalability by using a distributed cloud-based implementation.







\section*{Acknowledgment}

We acknowledge the use of ChatGPT and Microsoft Copilot for portions of code creation.


\bibliography{references}
\bibliographystyle{ieeetr}



\end{document}